\documentclass{article}
\usepackage{amssymb}
\usepackage[bookmarks=true]{hyperref}
\usepackage{amsfonts}
\usepackage{amsmath}
\usepackage{bm}
\usepackage{cleveref}
\usepackage{graphicx}
\usepackage{wrapfig}
\usepackage{multirow}
\usepackage{soul}

\usepackage[usenames, dvipsnames, table]{xcolor}

\usepackage[small]{caption}

\usepackage[final]{corl_2025} 

\usepackage[markup=bfit, deletedmarkup=sout, authormarkup=superscript]{changes}
\definechangesauthor[name={ar}, color=PineGreen] {AR}
\definechangesauthor[name={ap}, color=Purple] {AP}

\definecolor{pastelred}{HTML}{FF746C}
\definecolor{pink}{HTML}{FF9E99}
\definecolor{green}{HTML}{006837}
\definecolor{midgreen}{HTML}{84CA66}
\definecolor{mid}{HTML}{EFBF04}
\definecolor{midred}{HTML}{F98E52}
\definecolor{red}{HTML}{A50026}

\title{Dynamics-Compliant Trajectory Diffusion for Super-Nominal Payload Manipulation}

\author{
  Anuj Pasricha, Joewie Koh, Jay Vakil, Alessandro Roncone\\
  Department of Computer Science\\
  University of Colorado Boulder\\
  \texttt{firstname.lastname@colorado.edu}
}

\newcommand{\insertfig}{\vspace{-0.4in}
\includegraphics[width=\textwidth]{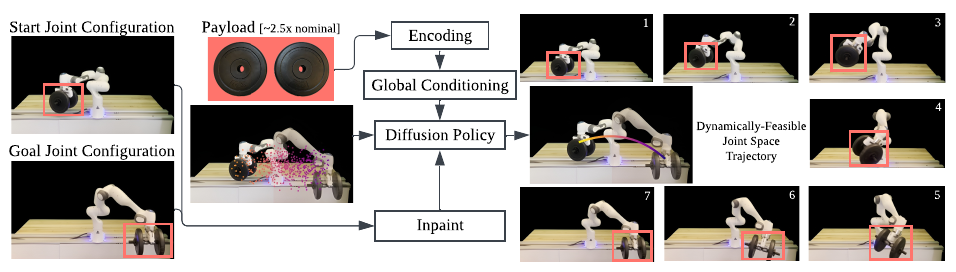}\captionof{figure}{The diffusion model presented in this work learns to generate dynamically feasible trajectories directly in joint angle, velocity, and acceleration space, enabling super-nominal payload manipulation.}\label{fig:firstpg}}

\apptocmd{\keywords}{\par\medskip{\centering\insertfig\par}}{}{}

\begin{document}
\maketitle

\begin{abstract}
Nominal payload ratings for articulated robots are typically derived from worst-case configurations, resulting in uniform payload constraints across the entire workspace. This conservative approach severely underutilizes the robot's inherent capabilities---our analysis demonstrates that manipulators can safely handle payloads well above nominal capacity across broad regions of their workspace while staying within joint angle, velocity, acceleration, and torque limits. To address this gap between assumed and actual capability, we propose a novel trajectory generation approach using denoising diffusion models that explicitly incorporates payload constraints into the planning process. Unlike traditional sampling-based methods that rely on inefficient trial-and-error, optimization-based methods that are prohibitively slow, or kinodynamic planners that struggle with problem dimensionality, our approach generates dynamically feasible joint-space trajectories in constant time that can be directly executed on physical hardware without post-processing. Experimental validation on a 7 DoF Franka Emika Panda robot demonstrates that up to $67.6\%$ of the workspace remains accessible even with payloads exceeding $3$ times the nominal capacity. This expanded operational envelope highlights the importance of a more nuanced consideration of payload dynamics in motion planning algorithms.
\end{abstract}

\keywords{Diffusion Models, Dynamics-Informed Planning, Payload Transport}

\section{Introduction}

Manipulator design establishes fundamental performance boundaries through the interplay of mechanical structure, actuation system, and control architecture.
At the hardware level, these capabilities manifest as explicit physical constraints: joint angle limits that define the reachable workspace, maximum velocities bounded by motor specifications, and torque limits determined by the drivetrain capacity.
However, a crucial distinction exists between these hardware limitations and the operational capacities specified by manufacturers for general-purpose deployment.
While hardware limits are absolute physical constraints, nominal specifications---like payload capacity and effective workspace boundaries---are \textit{derived} conservatively by considering worst-case scenarios across possible configurations and trajectories.
In practice, system integrators and end-users are locked into treating these derived operational capacities as hard limits rather than contextual capabilities, constrained by warranty requirements to design processes well within these conservative bounds.
This well-established approach of \textit{over-provisioning} hardware to ensure ample safety buffers has system-wide cascading effects---from the need to procure oversized actuators and power supplies, to reinforcing mounting structures for added load tolerance, to expanding facility requirements to accommodate larger equipment.
For instance, when an application requires handling a $35$kg payload, end-users are forced to select a $50$kg-rated robot over a $30$kg-rated system, despite the latter potentially being capable of safely executing the task within broad regions of its operational space.

Importantly, it is possible to overcome the issue of hardware over-provisioning through software-based optimizations, which allow to expand the safe operational envelope of a fixed/given robot embodiment and enable it to perform beyond nominal payload specifications.
This requires unified handling of both geometric (i.e., kinematic) and dynamic constraints, as payload characteristics interact with joint configurations and higher-order derivatives to determine feasible motions.
Even if they have not solved this specific problem, prior approaches in this space have investigated the challenge from multiple perspectives: geometric planners with dynamics post-processing \cite{orthey2023sampling,berscheid2021jerk}, kinodynamic planning \cite{shome2021asymptotically}, optimization-based methods \cite{ichnowski2022gomp}, and control-based frameworks \cite{arrizabalaga2024slosh}. However, these methods face inherent trade-offs between design complexity, constraint handling, and computational efficiency; these become particularly critical when operating near system limits where the feasible solution space becomes highly constrained.
These limitations highlight the need for a method that balances constraint satisfaction with
generalizability---i.e., the ability to adapt to varying payloads, environmental conditions, and task requirements while maintaining consistent performance.

Of note, diffusion models \cite{chi2023diffusionpolicy,sridhar2024nomad} present a compelling opportunity
address the ``curse of dimensionality'' inherent in trajectory planning with simultaneous kinematic and dynamic constraint satisfaction.
\textit{By learning from successful trajectories generated by multiple (imperfect) planning methods}, these models can extract various homotopic solution classes that overcome the limitations of individual approaches.
This distills successful solutions from diverse planning approaches into a unified representation that generalizes across the entire feasible space, including regions where individual planners fail.
The inherent stochasticity in the denoising process enables the exploration of diverse, multimodal trajectories \cite{chi2023diffusionpolicy}, while their capacity to model high-dimensional distributions allows incorporation of rich contextual information such as sensor data and problem constraints.
However, existing diffusion-based approaches have been confined to position-controlled systems \cite{carvalho2023motion,saha2024edmp}, leaving the fundamental challenge of integrating both kinematic and dynamic constraints unaddressed.

In this work, we propose a method that directly tackles this problem by jointly learning valid distributions of positions, velocities, and accelerations that satisfy both constraint types simultaneously---moving beyond prior work that only denoises in position space.
Through investigation of payload encoding strategies, we demonstrate that diffusion models effectively learn the complex relationships between joint states and payload-dependent dynamics, \uline{enabling constant-time generation ($\approx 10$ms) of feasible trajectories that preserve $67.6\%$ workspace accessibility at $3\times$ nominal payload capacity}---without explicit constraint checking at runtime. This demonstrates the importance of dynamics-informed generative models in expanding the operational boundaries of robotic systems.
\vspace{-0.3in}
\section{Related Work}
\vspace{-0.14in}
Manufacturer-specified operational parameters for general deployment of robots are typically far more conservative than hardware-level physical limits and often lack transparency in their derivation.
This has led to significant under-utilization of robotic capabilities, as evidenced by geometric motion planning and time parameterization approaches dominating most industrial use \cite{berscheid2021jerk,coleman2014reducing,orthey2023sampling} and recent advances in foundation models for manipulation in unstructured environments being restricted to exclusively geometric reasoning tasks \cite{o2024open,team2024octo,zitkovich2023rt}.
A nuanced, configuration-dependent understanding of the robot’s specifications that considers both robot and environment dynamics could enable a broader spectrum of manipulation capabilities within hardware limits.

Higher-order constraints such as joint torques and end-effector accelerations can be addressed through several approaches that range from ``plan-and-filter'' methods which post-process geometric paths \cite{abderezaei2024clutter,kuntz2019fast} to kinodynamic planners that directly incorporate dynamics during planning \cite{li2015sparse,nayak2022bidirectional,pasricha2024virtues}.
While these approaches offer completeness guarantees, they face practical limitations in high-dimensional manipulation spaces and often exhibit high variance in planning times especially when faced with state or model uncertainty.
These challenges are particularly acute when planning near operational limits, where the feasible solution space becomes increasingly constrained \cite{kingston2018sampling}.
Optimization-based methods naturally accommodate dynamics constraints but rely heavily on good initialization, leading to hybrid approaches that bootstrap the optimization process with learning or sampling-based solutions \cite{ichnowski2020deep,kuntz2019fast,yan2024impact}.
Optimal control formulations for dynamics-aware manipulation also exist \cite{arrizabalaga2024slosh,kim2014catching}. Despite their theoretical rigor, they face practical limitations in handling complex constraints and achieving reliable convergence for real-world manipulation problems.

Importantly, there exist a large body of works that demonstrate how a robot's behavioral repertoire can be significantly expanded when incorporating dynamic constraints---e.g. nonprehensile manipulation \cite{dogar2011framework,pasricha2022pokerrt,ruggiero2018nonprehensile,yu2016more,zeng2020tossingbot} or fast inertial transport \cite{arrizabalaga2024slosh,ichnowski2022gomp,muchacho2022solution,avigal2022gomp}.
These capabilities and considerations are particularly relevant for the specific problem of payload manipulation, where the interaction between payload and induced joint torques significantly influence task success.
Interestingly, payload manipulation presents unique challenges that intersect robot design and control.
Traditionally, prior work has focused on mechanical modeling for specific low-DoF systems \cite{wang2001payload}, with emphasis on payload capacity optimization \cite{korayem2008maximum} and joint torque considerations \cite{nho2003intelligent}. Other work in this domain tackles super-nominal payload handling through the use of multi-arm systems \cite{kim2021enhancing,yan2016coordinated}.
However, the field lacks generalizable approaches for high-DoF systems that can reason about dynamic payload-robot interactions across diverse tasks.

Learning-based approaches offer promising directions for both generalizability and efficient trajectory generation in high-dimensional constrained spaces while respecting system dynamics. Autoregressive models implicitly encode dynamics through dataset design \cite{fishman2023motion,qureshi2020motion}, while recent diffusion-based approaches provide non-autoregressive alternatives with improved sampling diversity \cite{carvalho2023motion,saha2024edmp}. Task-level diffusion policies, conditioned on language and visual embeddings \cite{ajay2023is,chi2023diffusionpolicy}, demonstrate potential for generating diverse behaviors from high-level task specifications.
Our method conditions a denoising diffusion model on the target payload to generate dynamically feasible trajectories.

\section{Diffusion Models for Trajectory Generation}\label{sec:diffusion}

Diffusion models are a class of generative models that prdouce high-quality and diverse samples across multiple data modalities \cite{ho2020denoising}. 
These models operate through a two-phase process: first, gradually corrupting training data with Gaussian noise in a forward process, then learning to reverse this corruption through an iterative denoising process.
In the context of robot motion planning, we use diffusion to generate collision-free and dynamically feasible trajectories.

A joint space trajectory with $t$ waypoints at denoising step $k$ is defined as
$\bm{\pi}^k = [ X_0^k, X_1^k, \dots, X_{t-1}^k ]^T$
where $X_i = (q_i, \dot{q}_i, \ddot{q}_i) \in \mathbb{R}^{3n}$ represents the complete state (joint angles, velocities, and accelerations) of an $n$-DoF robot arm at timestep $i$.
The training process for the diffusion model approximates the conditional distribution $p(\bm{\pi}\ |\ \mathcal{P})$, where $\mathcal{P}$ is the target payload representation (\Cref{sec:model}).
It does so by sampling $\bm{\pi}$ from dataset $\mathcal{D}$, adding noise according to a noise schedule parameterized by $\alpha$, $\gamma$, and $\sigma$, and learning to predict the underlying sample. The noise prediction network $\epsilon_\theta$, where $\theta$ represents the parameters of the 1D U-Net architecture, is trained to minimize
    $\mathcal{L}(\theta) = MSE(\epsilon^k, \epsilon_\theta(\mathcal{P}, \bm{\pi}^0 + \epsilon^k, k))$.

Given a trained $\epsilon_\theta$, the $K$-step iterative denoising process
    $\bm{\pi}^{k-1} = \alpha \cdot (\bm{\pi}^k - \gamma\epsilon_\theta(\mathcal{P}, \bm{\pi}^k, k) + \mathcal{N}(0, \sigma^2I))$
starts from $\bm{\pi}^K \sim \mathcal{N}(0, I)$ and generates a trajectory $\bm{\pi}^0$ that can support a target payload and obeys start, goal, joint limit, and collision constraints.
In practice, start and goal states are enforced via inpainting \cite{carvalho2023motion}, i.e., $\bm{\pi}_0^k = X_{start}$ and $\bm{\pi}_{t-1}^k = X_{goal}$ where $\dot{q}_{start} = \dot{q}_{goal} = \ddot{q}_{start} = \ddot{q}_{goal} = \bm{0}$.
Collisions are avoided via inference-time gradient guidance $\bm{\pi}^{k-1} = \bm{\pi}^k - \beta\ \cdot \nabla J(\bm{\pi}; E)$ where $\beta$ is a tunable guidance weight, $J$ defines the collision cost function, and $E$ represents the set of all obstacles in the environment \cite{saha2024edmp}.
Additionally, joint limits are enforced via clamping during the training and inference processes.
Specific to our work, $p(\bm{\pi}\ |\ \mathcal{P})$ implicitly models dynamically feasible payload-induced torques through an object property-based global conditioning mechanism that influences the entire trajectory by appending target payload representation $m$ to the diffusion timestep embedding in the noise prediction network architecture \cite{chi2023diffusionpolicy}.
\section{Payload-Conditioned Trajectory Denoising}

In this section, we present our payload-conditioned diffusion model to generate pick-and-place trajectories that transport super-nominal payloads while obeying kinematic and dynamic constraints.

\subsection{Training Data}

\begin{wrapfigure}{r}{0.5\textwidth}   
  \vspace{-1.2\baselineskip}            
  \centering
  \includegraphics[width=\linewidth]{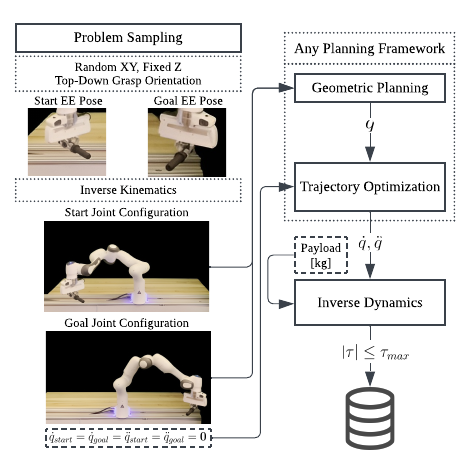}
  \caption{Plan-and-filter process to create training data.}
  \label{fig:data-generation}
  \vspace{-1.4\baselineskip}            
\end{wrapfigure}

Our payload-conditioned diffusion model (\Cref{sec:model}) is trained on $25,000$ time-parameterized, collision-free joint space trajectories (state dimension = DoF$\times3$, corresponding to joint angles, velocities, and accelerations) for tabletop pick-and-place motions. These trajectories are generated using a plan-and-filter pipeline that uses cuRobo, a GPU-accelerated trajectory optimization framework \cite{sundaralingam2023curobo}, followed by a dynamics validation step (\Cref{fig:data-generation}).
First, trajectories are generated for the 7DoF Franka Emika Panda robot arm with an effective payload mass of $0$kg.
Second, to ensure that payload-induced joint torques remain within hardware limits, each trajectory is verified through an analytically-derived dynamics model parameterized on real-world data \cite{gaz2019dynamic}.
This model captures both the standard dynamic elements (inertia, Coriolis, and gravity) and velocity-dependent joint friction.
By using a dynamics model fitted to real robot trajectory data, we ensure that our training dataset reflects the physical constraints observed in hardware.

This dynamics verification step utilizes a complete robot model where $q$, $\dot{q}$, $\ddot{q}$, and $\tau$ are $7\times1$ vectors representing joint angles, velocities, accelerations, and torques respectively.
The dynamics matrices include: $M(q)$ ($7\times7$ positive-definite, symmetric joint inertia matrix), $C(q,\dot{q})$ ($7\times7$ Coriolis/centrifugal matrix), $g(q)$ ($7\times1$ gravity compensation torque vector), and $f(\dot{q})$ ($7\times1$ joint friction torque vector). External task space forces $F_{ext}$ ($6\times1$) map to joint torques through the Jacobian $J(q)$ ($7\times7$).
The dynamics model is parameterized as:
\begin{equation}\label{eq:arm-dyns}
    \tau = M(q)\ddot{q} + C(q, \dot{q})\dot{q} + g(q) + f(\dot{q}) + J^{-1}(q)F_{ext}
\end{equation}
Note that this analytically-derived model represents the robot without external payloads which can affect model accuracy when considering varying payload masses in real-world deployment.
To account for payload-induced joint torques, we model the gravitational effects as a 6D Cartesian wrench transformed from the world frame to the end-effector frame.
This frame transformation is critical for accurately projecting gravitational forces acting on the payload along the robot's embodiment.
We represent the external gravitational wrench $\bm{F}_g \in \mathbb{R}^6$ in the world frame with a simplified point mass model:
$\bm{F}_g = mg \cdot [0,\ 0,\ -1,\ 0,\ 0,\ 0]^T$
where $m \in \mathbb{R}^+$ denotes payload mass and $g \approx 9.81 \text{ m/s}^2$ is gravitational acceleration. This representation assumes force application at the object's center of mass, neglecting potential moment contributions.
The wrench's structure combines linear forces (first three components) and angular forces (last three components), which are
mapped to induced joint torques through the manipulator Jacobian, contributing to the total joint torque vector (\cref{eq:arm-dyns}).

The training dataset $\mathcal{D}$ is constructed from these trajectories that satisfy both kinematic and dynamic constraints. Each element $d_i \in \mathcal{D}$ is a tuple of trajectory $\bm{\pi}_i$ and maximum supported payload $m_i$. 
Each $m_i$ is the maximum supported payload mass for trajectory $i$ subject to $|\tau(\bm{\pi}_i, m_i)| \leq \tau_{max}$, $\forall t \in [0,T]$ where $\tau_t$ is computed via the dynamics equation (\cref{eq:arm-dyns}).
Having established our dataset $\mathcal{D}$ of dynamically feasible trajectories and their corresponding maximum payload limits, we now consider different ways of incorporating payload conditioning into the diffusion model architecture.

\vspace{-0.12in}

\subsection{Payload-Conditional Trajectory Generation}\label{sec:model}
\vspace{-0.1in}

\begin{wrapfigure}{l}{0.48\textwidth}   
  \vspace{-1.6\baselineskip}            
  \centering
  \includegraphics[width=\linewidth]{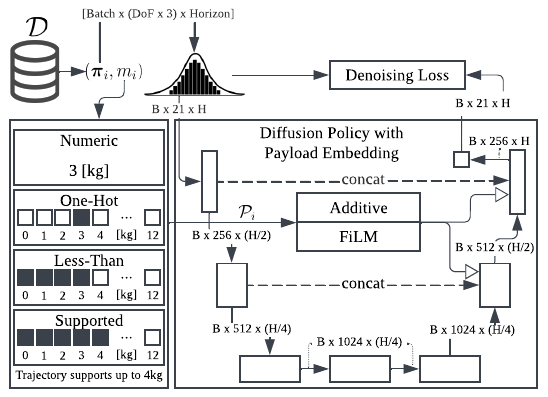}
  \caption{Our model conditions a 1D UNet denoising architecture \cite{chi2023diffusionpolicy} on various payload embeddings.}
  \label{fig:architecture}
  \vspace{-1.3\baselineskip}            
\end{wrapfigure}

We extend the diffusion policy architecture to condition trajectory generation on payload mass through global conditioning \cite{chi2023diffusionpolicy}. Unlike local conditioning where features influence each waypoint individually, global conditioning applies the mass embedding uniformly across the entire trajectory generation process, preserving dynamic feasibility across the generated motion (\Cref{fig:architecture}).
For each training trajectory $i$ with corresponding maximum supported payload mass $m_i \in \mathcal{D}$, we investigate four approaches to incorporate a target payload mass $p_i$ into an encoding vector $\mathcal{P}_i$.
The payload encoding $\mathcal{P}$ is concatenated with a diffusion timestep embedding to form a global conditioning vector, which is then integrated into $\epsilon_\theta$ through either additive or affine transform conditioning \cite{perez2018film}.
We consider the following approaches for designing $\mathcal{P}$.
\noindent\textbf{Numeric Input.} A random supported mass scalar $p_i \sim \mathcal{U}(0, m_i)$ is sampled and fed directly to the network. This approach naturally supports continuous payload values during inference, despite training on a discrete set of payload values.
\noindent\textbf{One-Hot Encoding.} For a system supporting payloads from $0-18$kg in $1$kg increments, a random supported mass $p_i \sim \mathcal{U}(0, m_i)$ is encoded as a $19$-dimensional binary vector with a single non-zero entry at index $\lceil p_i \rceil$.
While this categorical representation simplifies the learning task by discretizing the payload space, it constrains inference to the discrete payload values seen during training.
During inference, continuous payload masses must be quantized to their ceiling value $\lceil p \rceil$ before input to the denoising network, ensuring a tight upper bound for conservative operation.
\noindent\textbf{Less-Than Encoding.} This encoding exploits the logical constraint that trajectories supporting a given payload must also support all lighter payloads.
For a random supported mass $p_i \sim \mathcal{U}(0, m_i)$, we construct a $19$-dimensional binary vector where entries are $1$ for indices $j \leq \lceil p_i \rceil$ and $0$ otherwise.
This representation directly encodes the downward compatibility constraint into the input space, potentially improving the network's ability to learn and respect payload-dependent constraints.
During inference, continuous payload masses are similarly mapped to their ceiling value $\lceil p \rceil$ to construct the binary encoding.
\noindent\textbf{Supported-Range Encoding.}
This approach encodes the complete range of supported payloads for a given trajectory.
For training trajectory $i$ with maximum supported mass $m_i$, we construct a $19$-dimensional binary vector where entries are $1$ for indices $j \leq \lceil m_i \rceil$ and $0$ otherwise.
During training, this encoding preserves the full payload compatibility information from $\mathcal{D}$.
At inference time, for a target payload mass $p$, the encoding can be constructed either as a one-hot vector with a single non-zero entry at index $\lceil p \rceil$, or as a less-than encoding with ones for indices $j \leq \lceil p \rceil$.
This dual interpretation enables both trajectory generation for specific payloads and exploration of the learned manifold of payload-trajectory relationships.
Broadly, each encoding method presents trade-offs between generalization capability and training simplicity.
\vspace{-0.12in}
\section{Evaluation}
\vspace{-0.05in}
To validate our method, we focus on pick-and-place motions---a canonical industrial robotics requirement where super-nominal payload manipulation can significantly extend the operational envelope of robot arms while obeying strict hardware limits.\footnote{Please refer to the supplementary material for more details on compute hardware, problem domain, experimental setup, and baseline methods. Videos are available here: \url{https://payload-diffusion.github.io/}}
In our experiments, payloads were rigidly attached to the end-effector with a geometrically centered and uniform mass distribution. This was a deliberate experimental choice necessitated by the limited grasping force of our parallel gripper, ensuring repeatable execution across trials. Importantly, this choice is not a fundamental limitation of our method: as shown in the supplementary video, our approach is capable of handling non-rigidly attached objects and payloads with modest center-of-mass offsets, though some failure modes at higher payloads are also highlighted and explained there.
During execution, invalid trajectories generated by the diffusion model are simply not executed. For valid trajectories that fail due to unmodeled dynamics (e.g., payload oscillations), standard robot safety stops are triggered automatically. Our approach relies on the manufacturer's joint velocity controller for trajectory tracking, which provides sufficiently accurate state tracking for the tasks evaluated in this work.

\subsection{Comparison of Payload Encodings}

\begin{wrapfigure}{r}{0.6\textwidth}   
  \vspace{-1.2\baselineskip}            
  \centering
    \includegraphics[width=\linewidth]{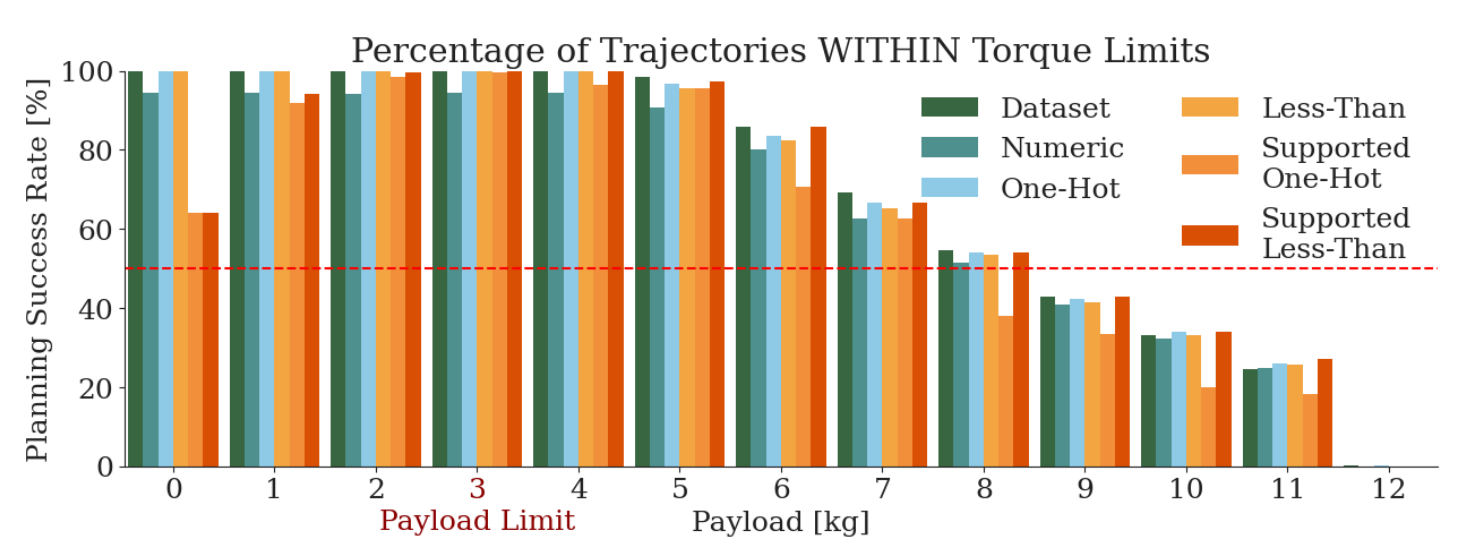}
  \caption{The aggregate success rate metric shows one-hot diffusion closely matching the underlying training data distribution.}\label{fig:bar}
  \vspace{-1.3\baselineskip}            
\end{wrapfigure}

Our evaluation of different payload encodings, conducted with FiLM conditioning and payload encoding normalization, reveals several key insights in terms of planning success rate (\Cref{fig:bar}).
Numeric and one-hot approaches demonstrate comparable effectiveness, particularly for payloads within nominal capacity ($0-3$kg). However, all methods show a tendency to underestimate success rates for higher payloads, potentially due to data imbalance in the training set. When interpreted as a less-than encoding during inference, the supported-range encoding achieves competitive performance, but its one-hot interpretation leads to significantly lower success rates.
While the encoding is designed to capture richer payload-trajectory relationships, the network struggles to fully utilize this information during generation.
The inability of numeric, supported one-hot, and supported less-than encodings to achieve $100\%$ success rates in the most loosely-constrained and data-rich portions of the state space, i.e., the nominal operating regime ($0-3$ kg), suggest fundamental limitations in how these encoding schemes enable the diffusion model to obey kinematic and dynamic constraints.
For subsequent experiments, we use with the one-hot encoding scheme, which demonstrates superior performance in terms of success rate.

\begin{figure}
  \centering
    \includegraphics[width=\columnwidth]{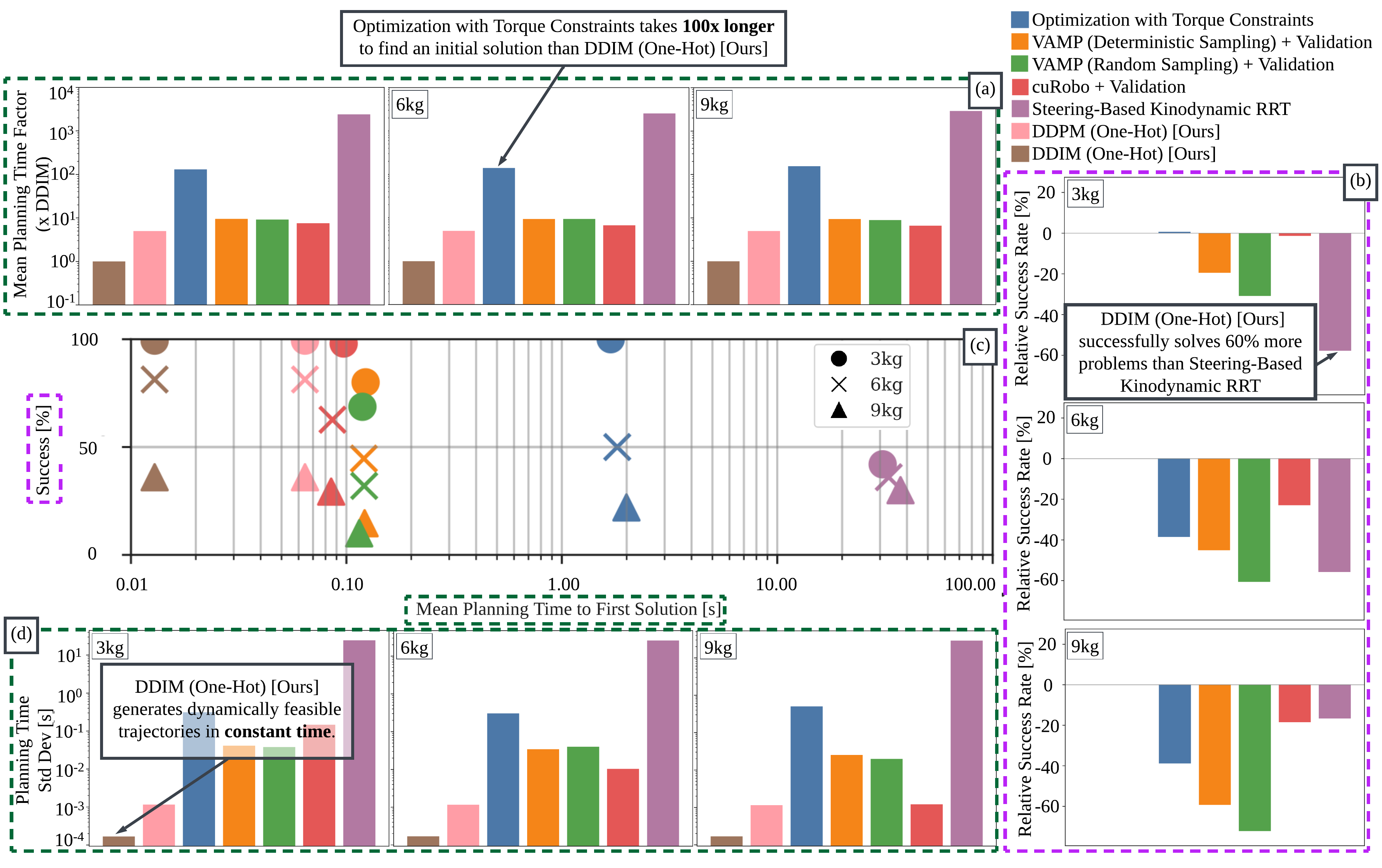}
  \caption{Comparative analysis of trajectory planning methods for payloads of $3$kg, $6$kg, and $9$kg. DDIM (One-Hot) significantly outperforms a variety of baseline methods in both planning time to first solution and success rate, generating dynamically feasible trajectories in constant time across all payload conditions.\vspace{-0.2in}}\label{fig:eval}
\end{figure}

\subsection{Comparison across Planner Types}

For the purposes of our discussion, we refer to DDIM (One-Hot) as a primary point of comparison. Denoising Diffusion Implicit Models (DDIM) offer accelerated inference due to fewer denoising steps (in our case, $5$) \cite{song2021denoising}. We also present results for DDPM \cite{ho2020denoising} with $25$ denoising steps. Notably, our experiments demonstrate that DDIM achieves identical success rates to DDPM despite requiring significantly less planning time ($\approx 0.01$s), validating our approach for realtime use (\Cref{fig:eval}c).

We evaluate our approach against a diverse set of trajectory planning methods using mean planning time to first solution and the corresponding success rate over $500$ tabletop manipulation tasks as the primary metric (\Cref{fig:eval}c). This metric provides a fair basis for comparison across fundamentally different planning paradigms. Additional metrics include planning time factor (mean planning time of baseline / mean planning time of DDIM) (\Cref{fig:eval}a), relative success rate change over DDIM as a percentage (\Cref{fig:eval}b), and standard deviations of planning times (\Cref{fig:eval}d). 

\paragraph{Plan-and-Filter Methods}

Broadly, plan-and-filter methods employ a sequential process: first planning a geometric path via sampling-based motion planning, then time-parameterizing it to derive joint velocities and accelerations, and finally validating trajectory feasibility for the target payload using the robot dynamics model (\cref{eq:arm-dyns}).
This iterative process continues until a feasible trajectory is found that satisfies all joint torque constraints induced by the payload \cite{abderezaei2024clutter}.
We evaluate two CPU-accelerated geometric planning variants: VAMP (RRTConnect) with deterministic sampling and VAMP (RRTConnect) with random sampling \cite{thomason2024motions}. We select Ruckig for time parameterization due to its ability to generate time-optimal trajectories in realtime \cite{berscheid2021jerk}.
Geometric motion planning and time parameterization can alternatively be unified within a single optimization framework (in our case, cuRobo) which can be followed by the dynamics validity checking.

Since there is a decoupling between kinematic reasoning and dynamic reasoning, further exacerbated by the separate time parameterization step in the VAMP variants, these approaches demonstrate inferior performance compared to our one-hot DDIM model (Fig. \ref{fig:eval}c).
VAMP with deterministic sampling achieves marginally higher success rates than VAMP with random sampling due to its low-dispersion sampling strategy, which provides more uniform exploration of the configuration space.
Despite this sampling advantage, both variants exhibit higher standard deviations (Fig. \ref{fig:eval}d) and lower overall success rates, particularly with heavier payloads (Fig. \ref{fig:eval}b). These methods also exhibit significant planning time variability (Fig. \ref{fig:eval}a).
The temporal predictability of our approach is particularly valuable for industrial pipelines, where high standard deviations in planning times can complicate integration with broader automation workflows.
The success rate differential between cuRobo and DDIM particularly demonstrates that our model effectively learns from successful trajectories generated by imperfect planners. While cuRobo with validation often requires multiple attempts to find feasible solutions, our method leverages these curated trajectories during training to achieve higher success rates, especially under super-nominal payload conditions.

\paragraph{Kinodynamic Motion Planning}
Plan-and-filter methods generate paths requiring post-processing to satisfy dynamic constraints, potentially invalidating the solution. By planning directly in the state space that includes joint angles, velocities and accelerations, kinodynamic planning ensures dynamically feasible solutions from the outset.
Specifically, we employ steering-based kinodynamic RRT as our baseline, planning in the joint angle, velocity, and acceleration space \cite{lavalle2001randomized}. Ruckig serves as our steering function owing to its ability to operate in realtime while respecting third-order constraints \cite{berscheid2021jerk}. Each edge undergoes validity checking that includes dynamics (\cref{eq:arm-dyns}) and smoothness checks before addition to the planning graph.
Results not only indicate high planning times to find the first solution ($>1000\times$ compared to DDIM) due to ``the curse of dimensionality'' (\Cref{fig:eval}a), but also show high standard deviations due to random sampling in the high-dimensional state space (\Cref{fig:eval}d).
The clustering of success rates observed across all three payloads suggests an inherent planning bias in kinodynamic RRT, indicating insufficient exploration in this highly non-Euclidean state space (\Cref{fig:eval}a).
Furthermore, kinodynamic approaches require extensive tuning of hyperparameters that interact with system dynamics in complex ways. Suboptimal parameter selection frequently results in planning timeouts or physically implausible trajectories.

\vspace{-0.08in}
\paragraph{Trajectory Optimization with Torque Constraints}

This baseline implements Sequential Least SQuares Programming (SLSQP) optimization for trajectory generation, minimizing sum-squared jerk to ensure smoothness \cite{ichnowski2022gomp}. To facilitate direct comparison with our method, we constrain trajectory durations instead of performing time optimization \cite{sundaralingam2023curobo,ichnowski2022gomp}, isolating evaluation to payload manipulation capabilities. The formulation incorporates linearized representations of both dynamic payload constraints and collision avoidance parameters.
The success rate disparity between optimization and DDIM becomes particularly pronounced with higher payloads as the optimization landscape becomes increasingly constrained.
This significant performance gap warrants discussion in the context of our evaluation choices. We implemented a basic SLSQP optimization formulation to establish a baseline that directly handles dynamic constraints without sophisticated warm-starting or acceleration techniques. Recent work has shown that optimization-based planners can achieve substantially better performance both in planning time and success rate through strategies like learning-based seeding \cite{ichnowski2020deep,huang2024diffusionseeder}.
However, our baseline still highlights the computational challenge of finding feasible solutions in high-dimensional spaces with nonlinear dynamic constraints.
\begin{figure}
  \centering
    \includegraphics[width=\columnwidth]{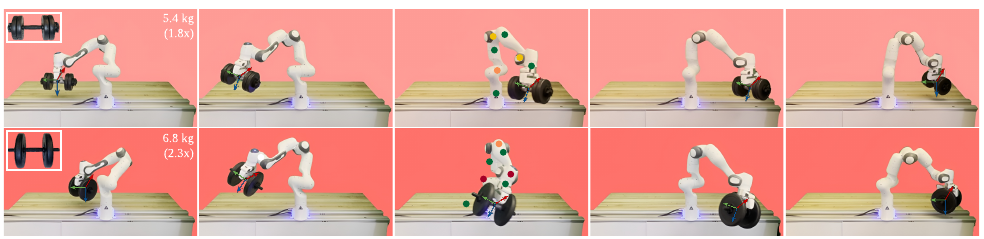}
  \caption{Two qualitative motions of the robot carrying super-nominal payloads of \textcolor{pink}{$5.4$kg ($1.8\times$ nominal capacity)} and \textcolor{pastelred}{$6.8$kg ($2.3\times$ nominal capacity)} are shown. Colored dots indicate joint torques as a percentage of joint limits: \textcolor{green}{$0-20\%$}, \textcolor{midgreen}{$20-40\%$}, \textcolor{mid}{$40-60\%$}, \textcolor{midred}{$60-80\%$}, \textcolor{red}{$80-100\%$}. Joints 1, 3, and 5 exhibit higher torques, with joint 1 bearing the structural load of the arm and joints 3 and 5 positioning and orienting the payload.\vspace{-0.2in}}\label{fig:vignette}
\end{figure}

\section{Conclusion}

This work demonstrates that manipulators can safely operate well beyond their nominal payload ratings through diffusion-based trajectory generation, challenging the traditional approach of hardware over-provisioning in industrial automation.
By training on trajectories that span the full range of payload-dependent dynamic constraints and leveraging the successes of ``imperfect'' planners, our model learns to generate motions for complex tasks in constant time that naturally respect system limitations without explicit constraint checking at runtime. This enables safe operation even at super-nominal payloads, maintaining $67.6\%$ of the nominal workspace at $3\times$ rated capacity.\footnote{More details on accessible workspace regions in the supplementary material.}

Potential avenues for future work include studying how to determine the optimal composition of training data from different planners operating in varied environments.
Moreover, the notion of conditioning on object-level characteristics can be generalized to handle objects with internal dynamics like liquids or articulated parts, moving beyond rigid-body assumptions. 
Such a framework could significantly reduce robotic system integration complexity and design burden by eliminating separate subsystems for different planning aspects, while maintaining the computational efficiency advantages and higher-order constraint satisfaction demonstrated in this work.

\clearpage
\section{Limitations}

While our results establish the viability of super-nominal payload manipulation on fixed/given hardware, several aspects warrant further discussion. \textbf{Rigid Attachment Choice.} Even though the rigid attachment experimental choice simplifies analysis by eliminating offset moments, it introduces two critical limitations in real-world execution (see accompanying video): dynamic effects from payload oscillation when objects are not perfectly rigid, and unmodeled torques when the object's center of mass is significantly offset from the end-effector's origin. These assumptions can lead to trajectory tracking errors or even task failures, particularly when handling flexible materials or asymmetric objects. Looking ahead, extending our approach to objects with internal dynamics will require explicit robot-object dynamics models to capture effects such as rotational inertia, slippage, and uneven mass distributions. Incorporating these nonlinearities, together with friction modeling for fingered grippers and suction force modeling for suction-based end-effectors, remains an important direction. Moreover, expanding the conditioning space to whole-body contact forces, fingertip wrench cones, and suction dynamics, as well as embedding more advanced controllers (e.g., impedance or admittance), will allow the same ``constraint-aware by design'' principle to scale to interaction-rich manipulation tasks. \textbf{Test-Time Robustness.} Assessing test-time robustness requires long-term deployment studies and, in many cases, proprietary Original Equipment Manufacturer (OEM) data on wear-and-tear. Whether robots should operate continuously at or near maximum capacity is also a design question for robot OEMs. While our method generates trajectories that respect manufacturer-specified torque limits, joints do run hotter under high payloads, motivating further study of safety margins under sustained operation. \textbf{Embodiment Generalization.} We acknowledge that embodiment understanding must be more directly incorporated into the training process to enable generalization across different robotic platforms. To this end, we are developing more general embodiment-constrained conditioning mechanisms as part of ongoing work.

Our work also raises important theoretical questions about completeness and optimality guarantees for learning-based motion planning. Even though we can argue that diffusion models trained on optimal trajectories will generate similarly optimal solutions, establishing formal completeness guarantees remains an open challenge.
Moreover, while our method demonstrates advantages in inference speed, traditional optimization-based approaches offer clearer constraint satisfaction guarantees and greater flexibility in adapting to new constraints without retraining. That said, training is not a bottleneck: we generate training trajectories across multiple payload conditions, with each trajectory verified for kinematic and dynamic feasibility throughout the robot's operational workspace. The compact dimensionality of our denoised samples enables fast retraining of the UNet and efficient creation of a high-quality dataset with comprehensive operational space coverage, but without the need for real-world payload manipulation data (assuming the existence of a robot dynamics model, which also lends to system modularity).

More rigorous evaluations comparing CPU versus GPU performance and Python versus C++ implementations would enable direct baseline comparisons but standardizing these benchmarks remains challenging. Our evaluation of baseline methods maintains industry-standard implementations of baselines without additional optimizations for deployment infrastructure.
Several other under-explored areas of research remain: optimal training data selection for maximizing planner generalization, disambiguating data imbalance from model generalization to explain the high variance in torques for higher payloads, leveraging diffusion models' continual learning capabilities at industrial scale, and conditioning on real-world sensor modalities like point clouds for deployment.

\acknowledgments{We thank the anonymous reviewers for their insightful feedback and constructive questions, which greatly improved the final version of this paper. This work was supported by the Office of Naval Research under Grant N00014-22-1-2482.}


\bibliography{example}  

\end{document}